\documentclass[10pt,conference]{IEEEtran}
\IEEEoverridecommandlockouts
\usepackage[utf8]{inputenc}
\usepackage{graphicx}
\usepackage{amsmath,amsfonts,amssymb}
\usepackage{url}
\usepackage{cite} 
\usepackage[T1]{fontenc}
\usepackage{hyperref}
\usepackage{pifont}
\usepackage[nolist,nohyperlinks]{acronym}
\usepackage{tikz}
\usetikzlibrary{arrows.meta, positioning, shapes.geometric, fit}
\begin{acronym}

\acro{1D}{one-dimensional}

\acro{2D}{two-dimensional}

\acro{16-QAM}{16 - quadrature amplitude modulation}

\acro{64-QAM}{64 - quadrature amplitude modulation}

\acro{3GPP}{3rd.~generation partnership project}

\acro{3G}{third generation}

\acro{4G}{fourth Generation}

\acro{5G}{fifth Generation}

\acro{AD}[A/D]{analog-to-digital}

\acro{ADSL}{asynchronous digital subscriber line}

\acro{AGC}{automatic gain control}

\acro{AI}{artificial intelligence}

\acro{AM}{amplitude modulation}

\acro{AMC}{automatic modulation classification}

\acro{AP}{access point}

\acro{ASIC}{application-specific integrated circuit}

\acro{AWGN}{additive white gaussian noise}

\acro{BB}{base-band}

\acro{BC}{broadcast channel}

\acro{BD}{block diagonalization}

\acro{BER}{bit error rate}

\acro{BPSK}{binary phase shift keying}

\acro{BS}{base station}

\acro{CDF}{cumulative distribution function}

\acro{CDMA}{code division multiple access}

\acro{CN}{core network}

\acro{CNN}{convolutional neural networks}

\acro{CQI}{channel quality indicator}

\acro{CR}{cognitive radio}

\acro{CSI}{channel state information}

\acro{CSIR}{channel state information at the receiver}

\acro{CSIT}{channel state information at the transmitter}

\acro{CSMA/CA}{carrier sense multiple access with collision avoidance}

\acro{CWT}{continuous wavelet transform}

\acro{DA}[D/A]{digital-to-analog}

\acro{DFT}{discrete Fourier transform}

\acro{DL}{deep learning}

\acro{DoF}{degrees of freedom}

\acro{DPC}{dirty paper coding}

\acro{DSA}{dynamic spectrum access}

\acro{DSP}{digital signal processor}

\acro{EDGE}{enhanced data rates for \acs{GSM} evolution}

\acro{EE}{energy efficiency}

\acro{FDD}{frequency division duplexing}

\acro{FFT}{fast Fourier transform}

\acro{FM}{frequency modulation}

\acro{FPGA}{field-programmable gate array}

\acro{FTTH}{fiber-to-the-home}

\acro{GSM}{global system for mobile communications}

\acro{GPP}{general purpose processor}

\acro{HF}{high frequency}

\acro{HSPA}{high speed packet access}

\acro{IA}{interference alignment}

\acro{IC}{index coding}

\acro{ISTFT}{inverse short-time Fourier transform}

\acro{IDFT}{inverse discrete Fourier transform}

\acro{IEEE}{Institute of Electrical and Electronics Engineers}

\acro{IF}{intermediate frequency}

\acro{i.i.d.}{independent and identically distributed}

\acro{IoT}{internet of things}

\acro{IQ}{in-phase/quadrature}

\acro{IRBD}{iterative regularized block diagonalization}

\acro{ISM}{industrial, scientific and medical}

\acro{KKT}{Karush-Kuhn-Tucker}

\acro{LAN}{local area network}

\acro{LRMC}{low rank matrix completion}

\acro{LoS}{line of sight}

\acro{LS}{least squares}

\acro{LTE}{Long Term Evolution}

\acro{MAC}{multiple access channel}

\acro{MAE}{mean absolute error}

\acro{MCS}{modulation and coding scheme}

\acro{MF}{matched filter}

\acro{MIMO}{multiple input - multiple output}

\acro{MMSE}{minimum mean square error}

\acro{MSE}{mean square error}

\acro{MU-VFDM}{multi-user Vandermonde-subspace frequency division multiplexing}

\acro{MUE}{macro-cell user equipment}

\acro{NC}{network coding}

\acro{OFDM}{orthogonal frequency division multiplexing}

\acro{OFDMA}{orthogonal frequency division multiple access}

\acro{ORBF}{opportunistic random beamforming}

\acro{OTA}{over-the-air}

\acro{PAN}{personal area network}

\acro{PAPR}{peak to average power ratio}

\acro{PC}{personal computer}

\acro{PCS}{personal communications systems}

\acro{PHY}{physical layer}

\acro{PLL}{phase-locked loop}

\acro{PMR}{private mobile radio}

\acro{PoC}{proof of concept}

\acro{QoS}{quality of service}

\acro{QPSK}{quadrature phase shift keying}

\acro{R-CNN}{recursive-\ac{CNN}}

\acro{RAT}{radio access technology}
\acroplural{RAT}[RATs]{radio access technologies}

\acro{RF}{radio frequency}

\acro{RIBF}{regularized inverse beamforming}

\acro{RX}{receiver}

\acro{SE}{spectral efficiency}

\acro{SDR}{software defined radio}

\acro{SIC}{successive interference cancellation}

\acro{SIGMF}{Signal Metadata Format}

\acro{SINR}{signal to interference plus noise ratio}

\acro{SIR}{signal to interference ratio}

\acro{SMMSE}{successive minimum mean square error}

\acro{SNR}{signal to noise ratio}

\acro{SOTA}{state-of-the-art}

\acro{SR}{software radio}

\acro{STFT}{short-time Fourier transform}

\acro{SUE}{small-cell user equipment}

\acro{SUS-ZFBF}{semi-orthogonal user selection \ac{ZFBF}}

\acro{SVD}{singular value decomposition}

\acro{TDD}{time division duplexing}

\acro{TDMA}{time division multiple access}

\acro{TCP/IP}{transmission control protocol / internet protocol}

\acro{TF}{time-frequency}

\acro{TIM}{topological interference management}

\acro{TX}{transmitter}

\acro{UE}{user equipment}

\acro{USB}{universal serial bus}

\acro{USRP}{universal software radio peripheral}

\acro{UWB}{ultra wide band}

\acro{VFDM}{Vandermonde-subspace frequency division multiplexing}

\acro{WIFI}{wireless fidelity}

\acro{YOLO}{You Only Look Once}

\acro{ZF}{zero forcing}

\acro{ZFBF}{zero force beamforming}

\end{acronym}

\title{CorteXforge: Scenario-Based RF Dataset Generation on Real Testbeds}
\author{Andréa Joly, Leonardo S. Cardoso and Jean-Marie Gorce%
\thanks{This work was supported by the French Defence Innovation Agency (Agence de l’innovation de défense – AID), Ministry of Armed Forces. Experiments presented in this work were carried out using the SLICES-RI/CorteXlab testbed. (see http://www.cortexlab.fr).}%
\\
Inria, INSA Lyon, CITI, UR3720, 69621 Villeurbanne, France\\
Email: andrea.joly@inria.fr, leonardo.cardoso@insa-lyon.fr}

\begin{document}

\maketitle
\begin{abstract}
Deep learning for \ac{RF} signal analysis relies on large annotated datasets, yet synthetic datasets trade physical realism for controllability, while uncontrolled \acl{OTA} measurements offer realism but limited reproducibility. We present \textbf{CorteXforge}, a scenario-based framework for generating realistic and reproducible RF datasets on programmable \ac{SDR} testbeds. High-level \ac{RF} scenarios are automatically translated into synchronized experiments executed on the SLICES-RI/CorteXlab platform, producing fully annotated \ac{SIGMF} datasets that capture hardware impairments and propagation effects. We validate CorteXforge through automatic modulation classification using ModForge, an \ac{OTA} dataset generated with the framework. Cross-domain experiments reveal a synthetic-to-real performance gap and show that synthetic pretraining remains beneficial for adaptation to \ac{OTA} data. CorteXforge provides a practical foundation for reproducible \ac{RF} dataset generation and future research on domain generalization.
\end{abstract}


\section{Introduction}\label{sec:intro}
\IEEEPARstart{R}{ecent} advances in deep learning have led to significant progress in radio signal analysis tasks, including \acf{AMC}, spectrum monitoring, and device identification. However, these approaches rely on large annotated datasets, whose availability remains a major bottleneck in the \ac{RF} domain.

Existing \ac{RF} datasets can be categorized into synthetic and real-world measurements. Synthetic datasets, such as RadioML2016 \cite{OShea2016RadioML, conrad} and RadioML2018.01A \cite{ovethe}, provide scalable generation and precise control over signal parameters, but cannot fully capture hardware impairment and complex propagation effects, limiting real-world generalization \cite{chillet}, \cite{oshea2018over}.

On the other hand, several measurement-based datasets such as ORACLE \cite{sankhe2019oracle} and Electrosense \cite{RajendranCFBCGP17} naturally capture propagation and hardware effects, but offer limited control over acquisition conditions and signal parameters. This makes systematic parameter variation and reproducible ML benchmarking more difficult.

To address this trade-off, we propose CorteXforge, a framework that describes RF experiments as structured scenarios specifying radio behaviors and signal parameters over time. These scenarios are then executed on the large-scale programmable \ac{RF} testbed SLICES-RI/CorteXlab \cite{cortexlabDoc1} and produce annotated datasets following the SigMF standard \cite{sigmf}. This approach generates datasets that combine the controllability of synthetic data with reproducible \ac{OTA} conditions.

We validate the proposed framework through a series of experiments on modulation classification tasks, comparing models trained on scenario-generated \ac{OTA} datasets with those trained on conventional synthetic datasets. Experimental results highlight the limitations of purely synthetic benchmarks, quantify the synthetic-to-real domain gap, and demonstrate the value of controlled \ac{OTA} datasets for evaluating and adapting models under physical \ac{RF} conditions. 

The main contributions of this work are summarized as follows:

\begin{itemize}
    \item We propose \textbf{CorteXforge}, a scenario-based framework for generating realistic and reproducible \ac{RF} datasets. The framework automatically translates high-level wireless scenarios into synchronized \ac{OTA} \ac{SDR} experiments and exports fully annotated datasets following the SigMF standard.

    \item We introduce \textbf{ModForge}, a controlled \ac{OTA} modulation dataset generated with CorteXforge, containing 24 modulation classes and multiple \ac{SDR} configurations.
    
    \item We quantify the synthetic-to-real performance gap for \ac{AMC} and evaluate the benefit of synthetic pretraining for adaptation to realistic \ac{RF} data.
\end{itemize}

\section{Related work}\label{sec:related}
%The availability of annotated \ac{RF} datasets has played a key role in the development of deep learning approaches for radio. Existing datasets can be broadly divided into synthetic and real-world datasets, each with distinct advantages and limitations.

Synthetic datasets, such as RadioML2016.10A \cite{OShea2016RadioML, conrad}, RadioML2018.01A \cite{ovethe}, and Sig53 \cite{larsca}, are widely used in the literature due to their scalability and precise control over signal parameters. These datasets typically provide labeled IQ samples with annotations including modulation type and \ac{SNR}. While more recent datasets such as Sig53 incorporate significantly richer channel models and simulated RF impairments than earlier benchmarks, they remain entirely simulation-based. Consequently, they cannot fully reproduce the richness of hardware-specific distortions, synchronization inaccuracies, and propagation effects encountered in over-the-air acquisitions, which often limits the generalization of models to real-world deployments.

Measurement-based datasets, such as DroneRF \cite{dronerf2019}, ORACLE \cite{sankhe2019oracle} and Electrosense \cite{RajendranCFBCGP17}, provide measurements collected from physical radio environments. These datasets inherently include realistic channel conditions and hardware-induced distortions, making them valuable for tasks such as RF fingerprinting, device identification, and spectrum monitoring. However, they are typically constrained by limited parameter control, restricted diversity, and high acquisition costs. In many cases, the data collection process is optimized to specific applications, making it difficult to systematically explore variations in modulation schemes, interference patterns, or \ac{SNR}, or multi-objective learning.

%These two dataset generation paradigms can therefore be viewed as opposite ends of a spectrum. Simulation provides a high degree of controllability, scalability, and reproducibility, at the cost of an imperfect representation of physical \ac{RF} effects. However, measurements collected in uncontrolled real-world environments naturally capture hardware, propagation, and interference effects but offer less control over experimental conditions and are generally more difficult to reproduce and annotate consistently.

Recent works have highlighted the importance of incorporating real-world RF data for deep learning-based PHY design. For instance, DeepWiPHY \cite{zhang2020deepwiphydeeplearningbasedreceiver} proposes a deep learning-based \acf{OFDM} receiver trained on both synthetic and over-the-air collected datasets. While such approaches demonstrate the benefits of realistic data acquisition, they do not provide a generic and reusable framework for dataset generation with controlled experimental scenarios.

Table~\ref{tab:rf_datasets} summarizes representative RF datasets.

\begin{table*}[t]
    \centering
    \footnotesize
    \begin{tabular}{lcccccc}
    \hline
    Dataset & Type & Multi-signal & Representation & Annotation & Task \\
    \hline
    RadioML2016.10A \cite{conrad, OShea2016RadioML} & Synthetic &  & IQ & Modulation, SNR & AMC \\
    RadioML2018.01A \cite{ovethe} & Synthetic &  & IQ & Modulation, SNR & AMC \\
    Sig53 \cite{larsca} & Synthetic &  & IQ & Modulation, SNR & AMC \\
    CRML23 \cite{xing2024jointsignaldetectionautomatic} & Synthetic & \ding{51} & Spectrogram & \acs{TF} boxes, Modulation & Detection + AMC \\
    ORACLE \cite{sankhe2019oracle} & Measurement &  & IQ & Emitter ID & RF Fingerprinting \\
    DroneRF \cite{dronerf2019} & Measurement &  & IQ & Drone ID & Device identification \\
    Electrosense \cite{RajendranCFBCGP17} & Measurement & \ding{51} & PSD / IQ & Spectrum occupancy & Spectrum monitoring \\
    \hline
    \end{tabular}
    \vspace{2mm}
    \caption{Comparison of commonly used \ac{RF} machine learning datasets.}\label{tab:rf_datasets}
\end{table*}

%In this work, we propose CorteXforge between these two dataset-generation paradigms. By relying on a programmable \ac{RF} testbed, CorteXforge enables over-the-air data acquisition while maintaining control over the experimental conditions. Compared with purely synthetic datasets, the resulting signals naturally include hardware impairments and propagation effects from physical radio transmissions. At the same time, the controlled testbed environment allows signal parameters, radio configurations, and transmission scenarios to be precisely defined and experiments to be reproduced under similar conditions. 

%However, this approach represents a trade-off rather than a replacement for synthetic or real-world data acquisition. While controlled \ac{OTA} experiments provide greater physical realism than simulation, they do not capture the full diversity and unpredictability of uncontrolled real-world environments, such as heterogeneous interference sources, mobility, and varying propagation conditions. CorteXforge provides an intermediate approach, combining the controllability and reproducibility of experimental scenarios with the physical effects of \ac{OTA} acquisition, while retaining the inherent limitations of a controlled testbed environment.

CorteXforge complements these approaches by using programmable RF platforms as a dataset-generation infrastructure. Unlike simulation-based datasets, signals are physically transmitted and acquired through SDR hardware; unlike uncontrolled field measurements, experimental conditions and signal parameters remain explicitly defined by the scenario. The resulting datasets therefore target reproducible evaluation under controlled OTA conditions.

\section{CorteXforge framework}\label{sec:cortexforge}

%This section presents \textbf{CorteXforge}, a scenario-based framework for generating realistic and reproducible \ac{RF} datasets on programmable \ac{SDR} testbeds. The proposed framework provides an intermediate approach between fully synthetic dataset generation and conventional \ac{OTA} measurements by enabling reproducible physical \ac{RF} experiments with precise control over experimental parameters.

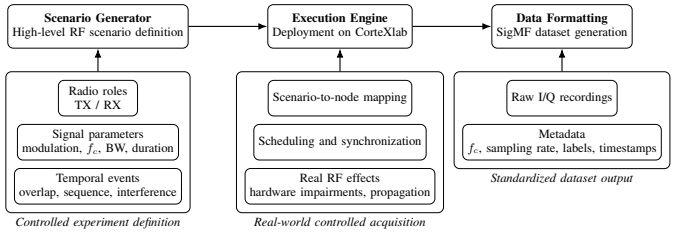
\begin{figure}[t]
    \centering
    \resizebox{\columnwidth}{!}{%
    \begin{tikzpicture}[
        font=\small,
        node distance=7mm and 8mm,
        >=Latex,
        box/.style={draw, rounded corners, align=center, minimum width=2.4cm, minimum height=1.1cm},
        smallbox/.style={draw, rounded corners, align=center, minimum width=2cm, minimum height=0.9cm},
        group/.style={draw, rounded corners, inner sep=6pt},
        arrow/.style={->, thick}
    ]

    % Main pipeline
    \node[box] (scenario) {\textbf{Scenario Generator}\\High-level RF scenario definition};
    \node[box, right=20mm of scenario] (engine) {\textbf{Execution Engine}\\Deployment on CorteXlab};
    \node[box, right=20mm of engine] (format) {\textbf{Data Formatting}\\SigMF dataset generation};

    \draw[arrow] (scenario) -- node[above, align=center] {} (engine);
    \draw[arrow] (engine) -- node[above, align=center] {} (format);

    % Scenario details
    \node[smallbox, below=8mm of scenario] (nodes) {Radio roles\\TX / RX};
    \node[smallbox, below=2mm of nodes] (params) {Signal parameters\\modulation, $f_c$, BW, duration};
    \node[smallbox, below=2mm of params] (events) {Temporal events\\overlap, sequence, interference};
    \node[group, fit=(nodes)(params)(events), label=below:{\textit{Controlled experiment definition}}](scenario1) {};

    \draw[arrow] (scenario1.north) -- ++(0,0.35) -| (scenario.south);

    % Execution details
    \node[smallbox, below=8mm of engine] (mapping) {Scenario-to-node mapping};
    \node[smallbox, below=2mm of mapping] (sched) {Scheduling and synchronization};
    \node[smallbox, below=2mm of sched] (rf) {Real RF effects\\hardware impairments, propagation};
    \node[group, fit=(mapping)(sched)(rf), label=below:{\textit{Real-world controlled acquisition}}](execution1) {};

    \draw[arrow] (execution1.north) -- ++(0,0.35) -| (engine.south);

    % Formatting details
    \node[smallbox, below=8mm of format] (iq) {Raw I/Q recordings};
    \node[smallbox, below=2mm of iq] (meta) {Metadata\\$f_c$, sampling rate, labels, timestamps};
    \node[group, fit=(iq)(meta), label=below:{\textit{Standardized dataset output}}](format1) {};

    \draw[arrow] (format1.north) -- ++(0,0.35) -| (format.south);

    \end{tikzpicture}
    }
    \caption{Overview of the CorteXforge pipeline.}
    \label{fig:system_overview}
\end{figure}

CorteXforge is a scenario-based framework that transforms high-level \ac{RF} experiment descriptions into synchronized \ac{OTA} acquisitions and annotated datasets on programmable \ac{SDR} testbeds. As illustrated in Fig.~\ref{fig:system_overview}, users describe the behavior of multiple radio nodes and their interactions over time, which the framework automatically translates into synchronized experiments on SLICES-RI/CorteXlab. 

The workflow consists of four successive stages, described in the following subsections:

\begin{enumerate}
    \item \ac{RF} scenario definition,
    \item scenario execution on the SLICES-RI/CorteXlab testbed,
    \item synchronized \ac{OTA} signal acquisition,
    \item automatic generation of annotated SigMF datasets.
\end{enumerate}

\subsection{RF Scenario Definition}

CorteXforge adopts a scenario-based approach to describe wireless experiments at a high level of abstraction. Instead of manually configuring individual \acp{SDR}, users specify the behavior of the radio environment through a structured scenario that defines the participating nodes, their roles (e.g., transmitter or receiver), and their transmission parameters (e.g., modulation scheme, transmission gain, carrier frequency, symbol rate, etc).

This high-level representation decouples experiment design from hardware configuration while providing fine-grained control over the generated radio environment. Rather than describing low-level \ac{SDR} commands, users only define the desired wireless scenario, which is subsequently translated into executable experiments by the framework.

Scenario files consist of one or more transmitting nodes and a single receiving node, with their role, and instructions over time. Receiver scenario file contains instructions mandatory for the receiver: sample rate, radio gain, carrier frequency, record duration. For each transmitting radio, scenario files specify temporal annotation, waveform characteristics (modulation scheme, signal amplitude, symbol rate), carrier frequency, transmission gain.

Temporal parameters allow transmissions to be precisely scheduled, making it possible to generate isolated signals as well as controlled interactions between multiple emitters, such as overlapping transmissions for interference patterns. Combined with the time synchronization provided by SLICES-RI/CorteXlab, this enables CorteXforge to automatically annotate precise temporal annotations for each transmission in the recorded IQ stream, without requiring manual labeling.

To generate diverse datasets, CorteXforge can explore the parameter space by varying the characteristics of successive scenarios. Typical parameters include the modulation format, transmission power, carrier frequency, symbol rate, number of simultaneous transmitters. This exploration enables the generation of datasets covering a broad range of controlled yet realistic RF conditions.

\subsection{Scenario Execution on SLICES-RI/CorteXlab Testbed}

CorteXforge is designed to be implemented on the SLICES-RI/CorteXlab platform, a large-scale programmable RF testbed designed for reproducible wireless experimentation \cite{cortexlabDoc1}. The platform consists of 40 \ac{SDR} nodes distributed inside a 180~m$^2$ electromagnetically shielded and partially anechoized room, enabling controlled \ac{OTA} experiments while limiting external radio interference and providing a stable propagation environment. Each radio node is equipped with a \ac{USRP} and an omnidirectional antenna, spaced at a regular interval of 1.8 m. A subset of the \ac{SDR} nodes shares common frequency and timing references distributed through OctoClock units, enabling simultaneous transmissions, precise time alignment, and accurate temporal annotations across synchronized nodes.

The execution engine maps the generated scenarios to physical radio nodes on the SLICES-RI/CorteXlab testbed. This step translates high-level scenario descriptions into low-level radio configurations and execution schedules.

Each radio node is configured with a GNU Radio flowgraph \cite{gnu-radio} according to the scenario specifications, including waveform generation, transmission parameters, and reception settings. 

Although the experimental environment is highly controlled, the acquired signals naturally include hardware impairments and environmental effects including oscillator inaccuracies, carrier frequency offsets (CFO), phase noise, amplifier nonlinearities, residual multipath propagation and antenna characteristics. Consequently, the collected datasets exhibit realistic \ac{RF} distortions that are difficult to reproduce comprehensively through simulation.

\subsection{Dataset Generation Pipeline}

The final stage of the pipeline consists in converting the acquired \ac{RF} data into a standardized and annotated dataset format.
Recorded signals are stored as complex \ac{IQ} samples, with 32-bit floating-point precision for both the in-phase and quadrature components, along with metadata describing the acquisition parameters. Labels are automatically derived from the scenario definitions, ensuring consistency between the generated signals and their annotations. This automated labeling process eliminates the need for manual annotation and ensures consistent annotations across repeated experiments.

In this framework, we adopt the SigMF (Signal Metadata Format) standard, which enables the storage of both raw signal samples (I/Q) and associated metadata.

For each record, the framework generates metadata describing key parameters:
\begin{itemize}
    \item Global parameters: sample rate, datatype, author
    \item Capture parameters: receiver radio characteristics such as gain, center frequency, hardware model
    \item Annotations: transmitter parameters such as modulation type, timing information (sample start and count), as well as additional physical-layer parameters such as symbol rate, roll-off factor, transmission gain, and transmitter-receiver distance
\end{itemize}

This standardized representation facilitates data sharing, reproducibility, and interoperability with existing signal processing and machine learning tools. 

The proposed pipeline produces fully annotated SigMF datasets that can subsequently be used for a wide range of RF machine learning tasks. In this work, it is used to generate the ModForge dataset introduced in Section~\ref{sec:experiment}.

\section{Experimental Protocol}\label{sec:experiment}

This section presents the experimental protocol used to evaluate the proposed framework. The objective is not to introduce a new modulation classification architecture, but rather to assess the impact of realistic over-the-air datasets on the generalization capabilities of deep learning models. To ensure a fair comparison, the same preprocessing pipeline, neural network architecture, and training procedure are used throughout all experiments.

\subsection{Datasets}

Two datasets are considered throughout this work.

The first dataset is RadioML2018.01A \cite{ovethe}, a widely adopted synthetic benchmark for \acs{AMC}. It contains \acs{IQ} recordings generated under controlled channel conditions, with annotations including modulation type and \acs{SNR}.

The second dataset, ModForge, is generated with CorteXforge on the SLICES-RI/CorteXlab testbed. Data acquisition involves four physical \acs{SDR} nodes, with three acting as transmitters and one as receiver during each acquisition. The receiver role is rotated across acquisitions such that each \ac{SDR} operates both as transmitter and receiver, and each modulation class is transmitted by every \ac{SDR}. All radios operate around 2.45~GHz, and the receiving node samples at 16.667~MHz. For each transmission, the receive \ac{SNR} is estimated within the expected occupied bandwidth derived from the waveform parameters as 
\begin{equation}
\mathrm{SNR}=10\log_{10}\!\left(\frac{P_{\mathrm{S+N}}-P_{\mathrm{N}}}{P_{\mathrm{N}}}\right),
\end{equation}
where $P_{\mathrm{S+N}}$ and $P_{\mathrm{N}}$ denote the in-band powers measured during the transmission and a noise-only reference interval, respectively. The resulting dataset contains 2,970,079 \ac{IQ} vectors of 1024 complex samples covering the same 24 modulation classes as RadioML2018.01A.

Table~\ref{tab:datasets} summarizes the properties of the two datasets used for this work.

\begin{table}[!t]
\centering
\caption{Comparison between RadioML2018.01A and the ModForge dataset.}
\label{tab:datasets}
\footnotesize
\begin{tabular}{lcc}
\hline
\textbf{Property} & \textbf{RadioML2018.01A} & \textbf{ModForge} \\
\hline
Acquisition & Synthetic & OTA (CorteXlab) \\
Representation & Complex IQ & Complex IQ \\
Window length & 1024 & 1024 \\
Modulation classes & 24 & 24 \\
SNR range & [-20, 30] & [-20, 30] \\
Hardware impairments & \ding{55} & \checkmark \\
Real RF propagation & \ding{55} & \checkmark \\
Multiple TX--RX & \ding{55} & \checkmark \\
\hline
\end{tabular}
\end{table}

\subsection{Neural Network Architecture}

To isolate the impact of the dataset from the impact of the learning algorithm, all experiments rely on the same \acs{CNN} architecture. The selected model is representative of commonly used \acs{AMC} architectures.

The network receives normalized \ac{IQ} samples as input and outputs one probability for each modulation class.

Models were implemented in PyTorch and trained using mixed-precision on an NVIDIA RTX 4000 Ada GPU. RadioML is split into training (70\%), validation (15\%), and test (15\%) subsets using stratified sampling with a fixed seed of 42. For ModForge, one of the four physical receivers is exclusively reserved for testing, while samples acquired by the remaining three receivers are split into training (85\%) and validation (15\%) subsets using modulation- and SNR-stratified sampling. This receiver-disjoint test protocol ensures that the receiving RF chain used for evaluation is never observed during training or model selection. However, all receivers operate within the same controlled CorteXlab environment; the split therefore provides receiver-disjoint evaluation rather than an independent propagation or deployment domain.

\begin{table}[!t]
\centering
\caption{CNN architecture used throughout the experiments.}
\label{tab:cnn}
\begin{tabular}{ll}
\hline
Component & Configuration \\
\hline
Input & IQ ($2\times1024$) \\
Feature extractor & 5 ConvBlocks \\
ConvBlock & 2 Conv1D + BN + GELU \\
Pooling & MaxPool1D \\
Global pooling & AvgPool + MaxPool \\
Classifier & 1024-512-256-24 \\
Activation & GELU \\
\hline
\end{tabular}
\end{table}

Table~\ref{tab:training_parameters} summarizes the training hyperparameters used throughout this work.

\begin{table}[]
    \centering
    \caption{Training configuration.}
    \label{tab:training_parameters}
    \begin{tabular}{lc}
    \hline
    Parameter & Value \\
    \hline
    Input shape & $2\times1024$ \\
    Output classes & 24 \\
    Optimizer & AdamW \\
    Initial learning rate & $2\times10^{-3}$ \\
    Weight decay & $10^{-4}$ \\
    Batch size & 1024 \\
    Maximum epochs & 100 \\
    Loss function & Cross-Entropy \\
    Input normalization & Per-sample normalization \\
    Learning rate scheduler & Step decay ($\times0.5$) \\
    Early stopping & Patience = 12 epochs \\
    Trainable parameters & 4.90 M \\
    \hline
    \end{tabular}
\end{table}

\subsection{Evaluation Protocol}

The objective of the experiments is to quantify the generalization capability of models trained on synthetic datasets when evaluated on realistic RF acquisitions.

Three evaluation scenarios are considered.

\begin{enumerate}

\item \textbf{In-domain evaluation.}

The model is trained and evaluated on RadioML using the conventional train/validation/test split. This experiment establishes the reference performance commonly reported in the literature.

\item \textbf{Cross-domain evaluation.}

The model trained on RadioML is directly evaluated on the CorteXforge dataset without any additional optimization. This experiment measures the synthetic-to-real generalization gap.

\item \textbf{Domain adaptation.}

Finally, the RadioML-pretrained model is fine-tuned using increasing fractions of the ModForge training set and evaluated on the receiver-disjoint ModForge test set. This experiment measures the benefit of controlled OTA data for domain adaptation while preventing the test receiver \ac{RF} chain from being observed during fine-tuning.

\end{enumerate}

\section{Results}\label{sec:results}

\subsection{In-domain Evaluation}

To establish a reference performance, the proposed CNN is first trained and evaluated on the RadioML2018.01A benchmark following the protocol described in Section~\ref{sec:experiment}.

\begin{figure}[!t]
\centering
\includegraphics[width=0.95\linewidth]{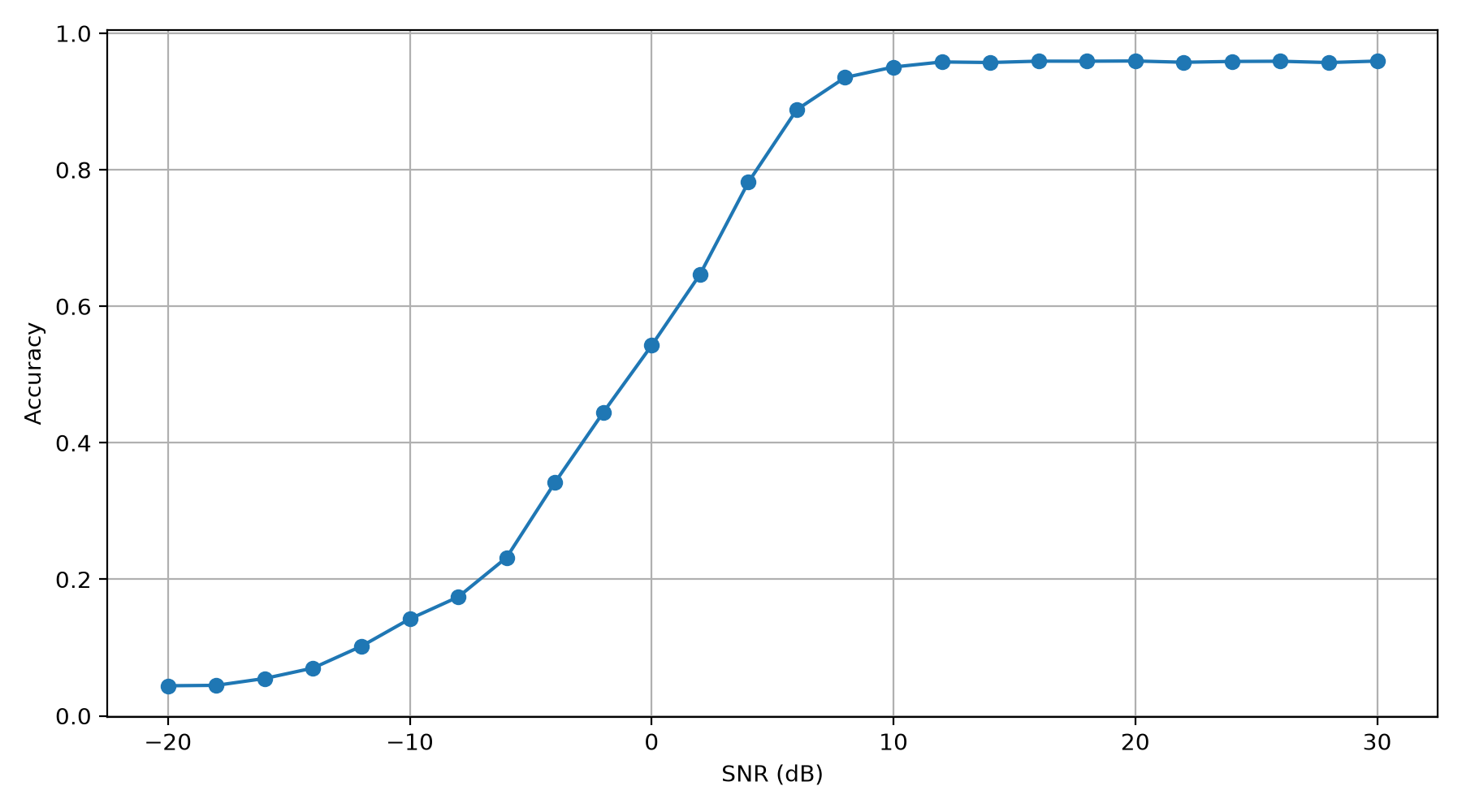}
\caption{Accuracy versus \acs{SNR} on the RadioML2018.01A test set.}
\label{fig:radioml_snr}
\end{figure}

Figure~\ref{fig:radioml_snr} reports the classification accuracy as a function of the \acs{SNR}. As expected, the model achieves high performance under synthetic conditions, with accuracy increasing with the \acs{SNR}. Classification errors occur mainly at low \acs{SNR}, where noise severely degrades the signal structure. Under these conditions, the classifier tends to predict a small subset of modulation classes, reflecting the limited discriminative information available in the input signal.

%Figure~\ref{fig:radioml_cm} presents the corresponding confusion matrix obtained on the RadioML test set. 

%\begin{figure}[!t]
%\centering
%\includegraphics[width=\linewidth]{figures/radioml_confusion_matrix.pdf}
%\caption{Confusion matrix obtained on the RadioML2018.01A test set.}
%\label{fig:radioml_cm}
%\end{figure}

Overall, the proposed model achieves 61.4\% average accuracy across the complete \ac{SNR} range, with accuracy exceeding 90\% at high \ac{SNR}. This performance is comparable to \cite{s25082489, s25165061}, providing a representative in-domain baseline for the subsequent cross-domain evaluation.

\subsection{Cross-domain Evaluation}

To evaluate the domain generalization capability of synthetic datasets, the \acs{CNN} trained exclusively on RadioML2018 was directly evaluated on the ModForge dataset without any retraining or fine-tuning. ModForge samples were processed using the same \ac{IQ} normalization as RadioML, and the trained network was applied without modifying its parameters. This protocol evaluates the performance degradation associated with the synthetic-to-\ac{OTA} domain shift.

\begin{figure}[!t]
\centering
\includegraphics[width=0.95\linewidth]{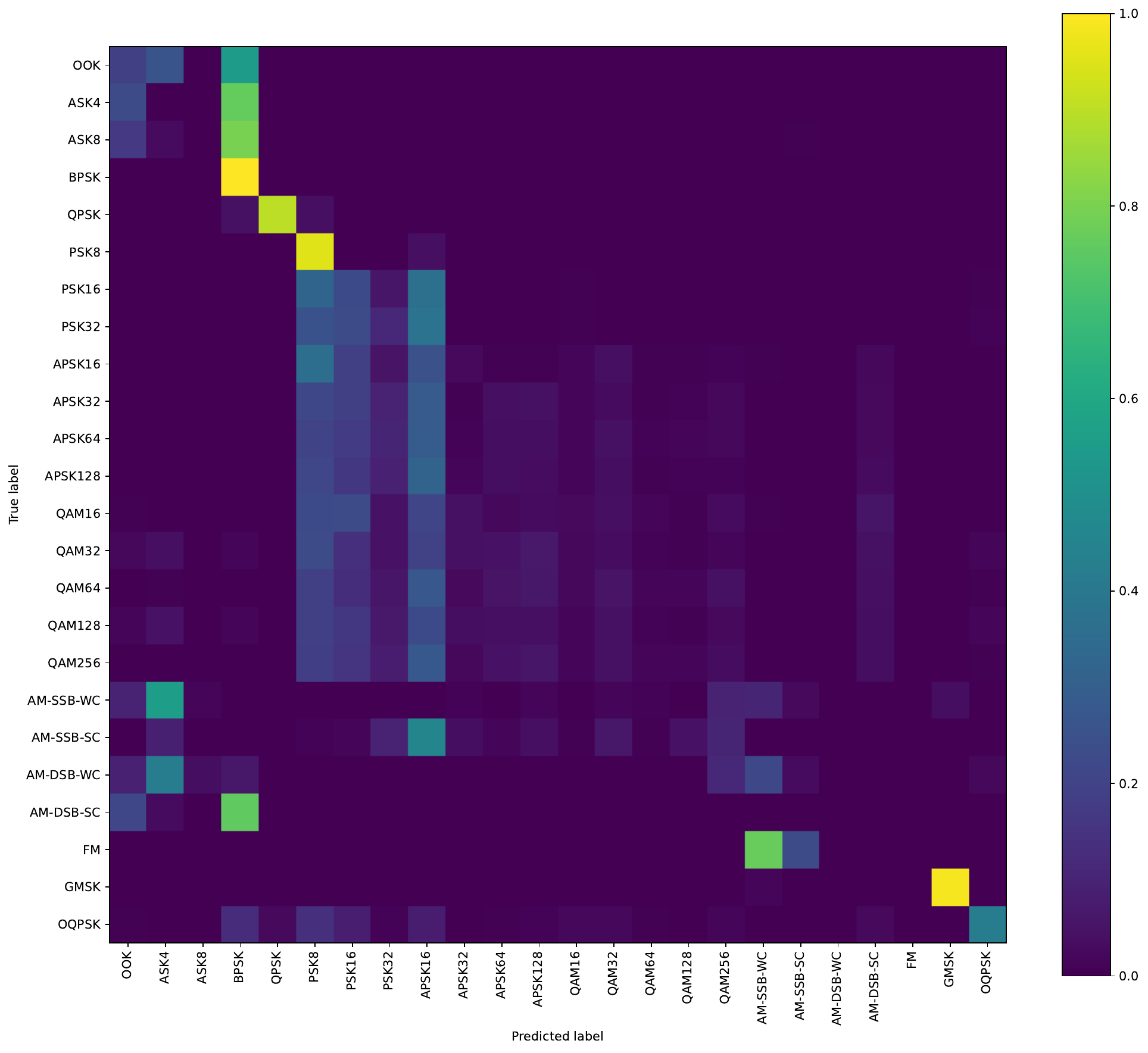}
\caption{Confusion matrix obtained on the ModForge dataset.}
\label{fig:cross_confusion}
\end{figure}

Figure~\ref{fig:cross_confusion} reports the resulting confusion matrix. Although the network still correctly recognizes a limited number of modulation families, the overall performance drops dramatically compared to the in-domain evaluation. Numerous modulation classes collapse towards a small subset of predicted classes, indicating that the learned feature representation does not transfer well to real \ac{OTA} recordings.

Global accuracy decreases from 61.4\% on the RadioML2018 test set to only 23.2\% on ModForge. This degradation highlights the large discrepancy between synthetic and real RF environments, despite both datasets containing the same modulation classes and identical \ac{IQ} representations.

\begin{figure}[!t]
\centering
\includegraphics[width=0.95\linewidth]{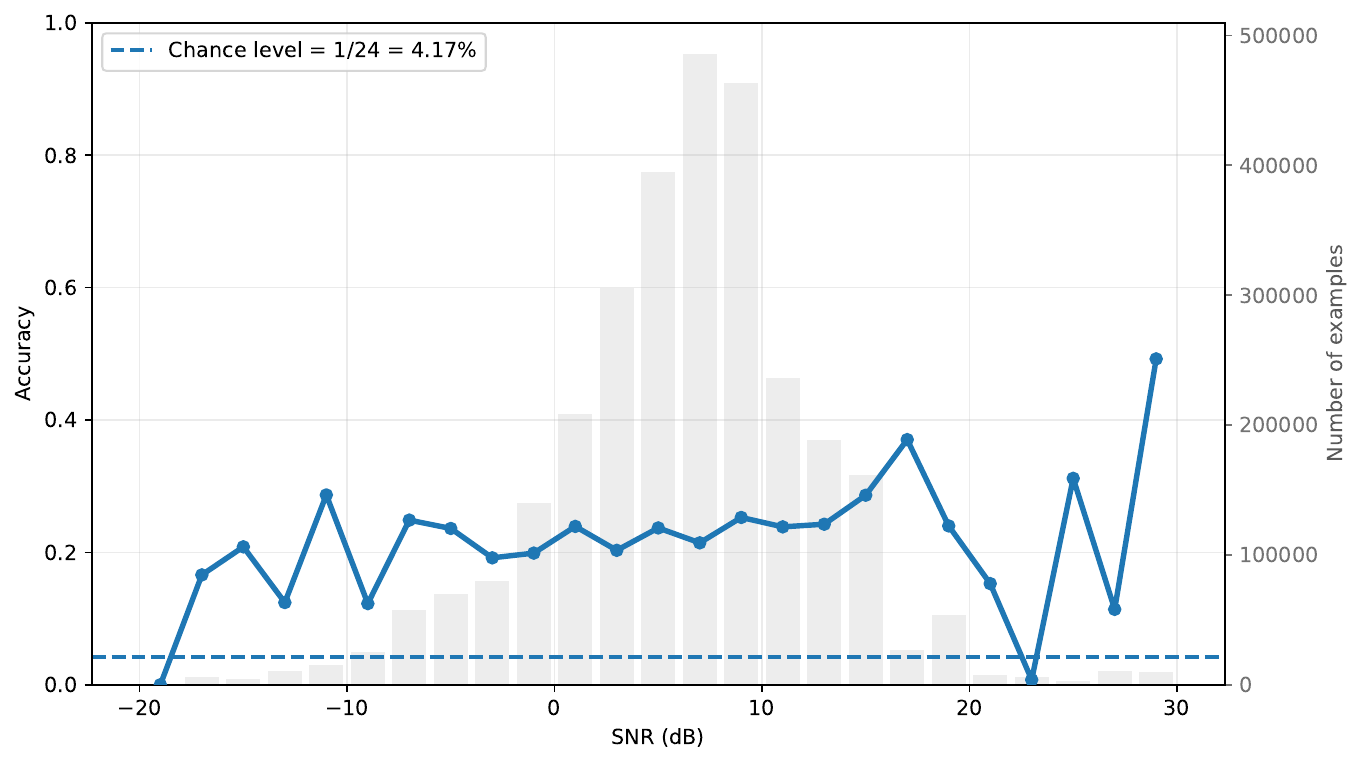}
\caption{Accuracy versus \acs{SNR} on the ModForge dataset.}
\label{fig:cross_snr}
\end{figure}

In Figure~\ref{fig:cross_snr} we see the classification accuracy as a function of the estimated received \acs{SNR}, with the blue plain line showing the accuracy as a function of the SNR (left vertical axis) and the gray bars the number of samples for each SNR range present in the set (right vertical axis). 

Unlike the in-domain evaluation, accuracy shows no clear monotonic improvement with increasing estimated receive \ac{SNR}. This suggests that additive noise alone cannot explain the observed degradation, which is largely associated with the synthetic-to-\ac{OTA} domain shift.

These results suggest that high in-domain accuracy obtained on synthetic datasets does not necessarily translate into good performance on physical \ac{OTA} recordings. Consequently, evaluating modulation classifiers exclusively on synthetic benchmarks may significantly overestimate their practical generalization capability.

\subsection{Transfer Learning}

Although the RadioML-trained model exhibits poor zero-shot generalization on ModForge, the learned representations may still provide a useful initialization for real-domain adaptation. To evaluate this hypothesis, we compare two training strategies. In the first one, referred to as \emph{from scratch}, the CNN is randomly initialized and trained only on ModForge. In the second one, referred to as \emph{RadioML-pretrained}, the CNN is first trained on RadioML2018 and then fine-tuned on ModForge.

Fine-tuning is performed using increasing fractions of the ModForge training set: 0.1\%, 0.5\%, 1\%, 5\%, 10\%, 25\%, 50\%, and 100\%. Training subsets are class-stratified and nested, such that each smaller subset is contained in the larger ones. The validation and receiver-disjoint test sets remain fixed across all experiments.

Both strategies use the same architecture, preprocessing pipeline, optimizer, and evaluation protocol, with a lower learning rate of $5\times10^{-5}$ used for fine-tuning.

\begin{figure}[!t]
\centering
\includegraphics[width=0.95\linewidth]{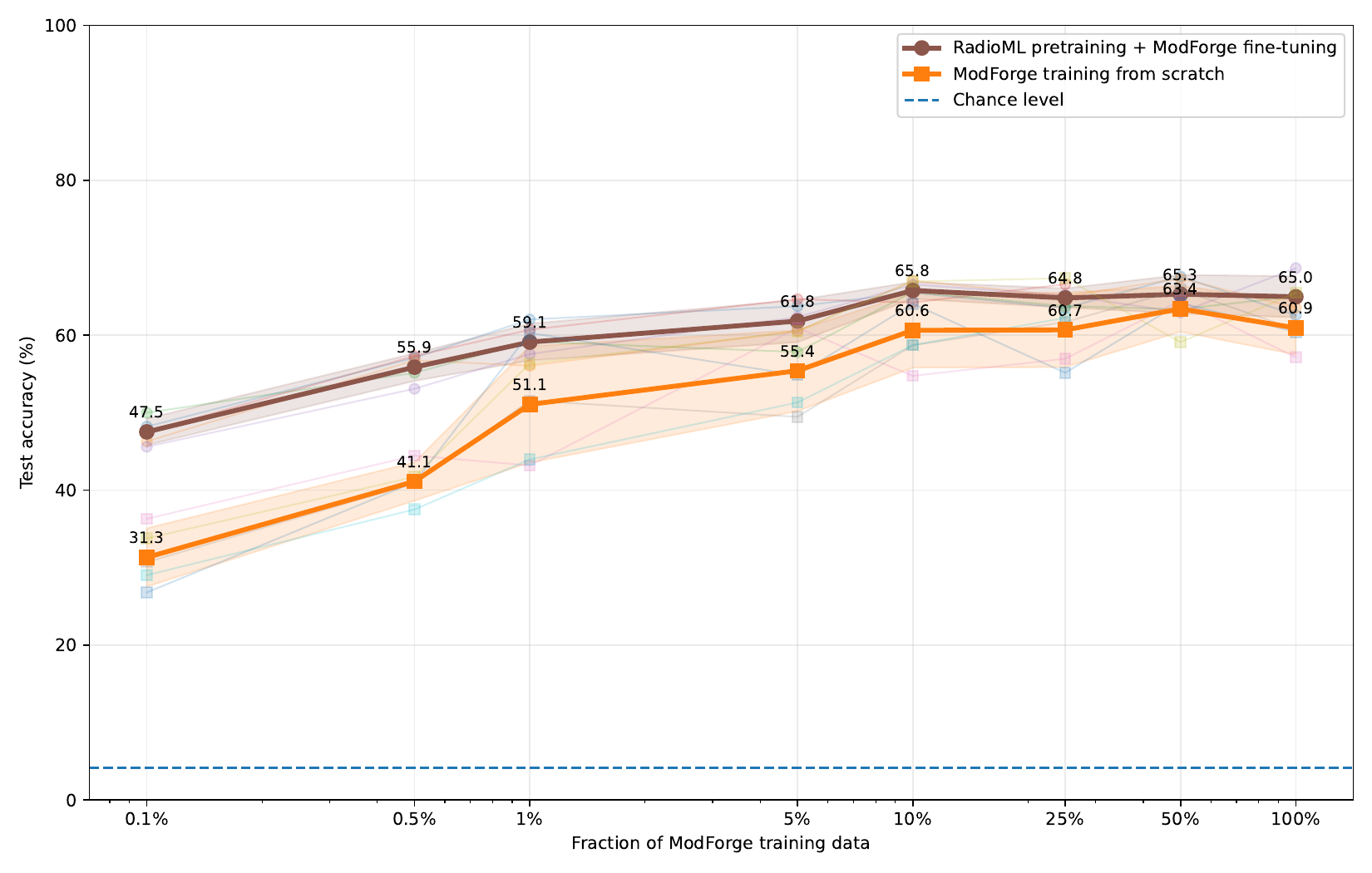}
\caption{Transfer learning performance on ModForge using increasing fractions of the training set. Curves report mean test accuracy across multiple random seeds}
\label{fig:transfer_learning}
\end{figure}

Figure~\ref{fig:transfer_learning} reports the test accuracy obtained on ModForge as a function of the fraction of \ac{OTA} training data. The \emph{RadioML-pretrained} model consistently matches or outperforms the model trained from scratch. The improvement is particularly visible in the low-data regime, where the \emph{RadioML-pretrained} model reaches higher accuracy with only a small fraction of ModForge samples.

These results show that synthetic pretraining is not sufficient for direct deployment on real \ac{OTA} acquisitions, as demonstrated by the zero-shot cross-domain evaluation. However, pretraining provides a useful initialization that facilitates adaptation to the real domain. This suggests that synthetic datasets and realistic \ac{OTA} datasets should be viewed as complementary rather than competing resources: synthetic data can be used for large-scale pretraining, while a limited amount of \ac{OTA} data can be used to adapt the model to realistic RF conditions. 

\section{Discussion}\label{sec:discussion}

\subsection{Role of Synthetic Data}

Our results highlight two complementary properties of synthetic RF data. First, the large zero-shot performance drop from RadioML to ModForge shows that in-domain synthetic performance alone is not a reliable proxy for performance on OTA acquisitions. Second, the transfer-learning experiments show that synthetic pretraining remains beneficial when adapting to the OTA domain, particularly when limited OTA training data are available. These observations suggest that synthetic and OTA datasets serve complementary roles: synthetic data provide scalable pretraining resources, while controlled OTA data enable evaluation and adaptation under physical RF conditions.

\subsection{Scope and Limitations}

CorteXforge should be viewed as complementary to both simulation and uncontrolled field measurements. Its purpose is not to reproduce the full diversity of real-world radio environments, but to provide controlled and reproducible physical acquisitions in which signal parameters, hardware configurations, and temporal interactions can be systematically varied.

Although ModForge includes multiple SDR front-ends and TX–RX configurations, all acquisitions were performed within the same SLICES-RI/CorteXlab environment, and the experimental validation was limited to modulation classification.

%Although demonstrated in this paper through automatic modulation classification, the proposed framework is designed as a generic RF scenario generation platform. The same workflow can be used to generate datasets for RF fingerprinting, localization, spectrum sensing, interference analysis, or future foundation models trained on realistic radio environments.

Moreover, CorteXlab is an isolated and controlled environment in which external interference is largely absent. As a result, ModForge does not capture the diversity of uncontrolled emitters, interference patterns, mobility, and propagation conditions encountered in open-spectrum deployments. Consequently, the present experiments characterize generalization across synthetic and controlled OTA domains, rather than generalization to arbitrary real-world environments. Future work will evaluate CorteXforge across additional RF tasks and environments and extend the scenario definitions to incorporate more diverse interference and propagation conditions

\section{Conclusion}\label{sec:conclusion}
In this paper, we introduced CorteXforge, a scenario-based framework for generating realistic, reproducible RF datasets on the SLICES-RI/CorteXlab testbed. Our experiments demonstrate that models trained on synthetic data (RadioML2018) drop when evaluated on \ac{OTA} acquisitions (ModForge), highlighting the critical need for realistic training data. By leveraging programmable testbeds, CorteXforge captures hardware impairments and propagation effects while maintaining controllability and reproducibility. Future work will explore scalability to larger datasets. The implementation of CorteXforge and the ModForge dataset are publicly available at: \url{https://github.com/Andreaj42/CorteXforge}.

\bibliographystyle{IEEEtran}
\bibliography{references}

\end{document}